\documentclass[11pt]{article}

\def\toicml{0}
\def\toacl{1}

\usepackage{acl2001}
\usepackage{graphicx}

\ifnum \toacl=1
\title{Using the Distribution of Performance for Studying Statistical
NLP Systems and Corpora}
\author{Yuval Krymolowski\\
        Department of Mathematics and Computer Science\\
        Bar-Ilan University\\
        52900 Ramat Gan, Israel\\
}

\fi

\newcommand\ignore[1]{}
\def\sn{{\rm SN}}

\begin{document}

\maketitle

\begin{abstract}
Statistical NLP systems are frequently evaluated and compared on the basis of their
performances on a single split of training and test data.
Results obtained using a single split are, however, subject to sampling noise. In this
paper we argue in favour of reporting a distribution of performance figures, obtained by
resampling the training data, rather than a single number.
The additional information from distributions can be used to make statistically quantified
statements about differences across parameter settings, systems, and corpora.
\end{abstract}

\section {Introduction}

The common practice in evaluating statistical NLP systems is using a standard
corpus (e.g., Penn TreeBank for parsing, Reuters for text categorization) along with
a standard split between training and test data.
As systems improve, it becomes harder to achieve additional
improvements, and the performance of various state-of-the-art systems is
approximately identical. This makes performance comparisons difficult.

In this paper, we argue in favour of studying the {\em distribution} of performance,
and present conclusions
drawn from studying the recall distribution.
This distribution provides measures for answering the following questions:
\begin{description}
\item[Q1:] Comparing systems on given data: Is classifier $A$ better than classifier
$B$ for given training and test data?
\item[Q2:] Adequacy of training data to test data: Is a system trained on dataset $X$
adequate for analysing dataset $Y$?
Are features from $X$ indicative in $Y$?
\item[Q3:] Comparing data sets with a given system: If a different training set improves the
result of system $A$ on dataset $Y_1$, will this be the case on dataset $Y_2$ as well? 
\end{description}
The answers to these questions can provide useful insight into statistical NLP systems.
In particular, about sensitivity to features in the training data, and transferability.
These properties can be different even when similar performance is reported.

A statistical treatment of Question 1 is presented by \newcite{yeh-signif}.
He tests for the significance of performance differences
on fixed training and test data sets. In other related works, \newcite{martin96small}
provides an overview of significance tests of error differences
in small samples, and \newcite{dietterich98approx} discusses results of a number of tests.

Questions 2 and 3 have been frequently raised in NLP, but not explicitly addressed,
since the prevailing evaluation methods provide no means of addressing them. In this
paper we propose addressing all three questions with a single experimental methodology,
which uses the distribution of recall.

\section{Motivation}

Words, parts-of-speech (POS), words, or any feature in text may be regarded as
outcomes of a statistical process. Therefore, word counts, count ratios, and other data
used in creating statistical NLP models are statistical quantities as well,
and as such prone to {\em sampling noise}. Sampling noise results from the finiteness
of the data, and the particular choice of training and test data.

A model is an approximation or a more abstract representation of training data.
One may look at a model as a collection of estimators analogous, e.g., to the slope
calculated by linear regression. These estimators are statistics with a distribution
related to the way they were obtained, which may be very complicated. The
performance figures, being dependent on these estimators, have
a distribution function which may be difficult to find theoretically.
This distribution gives rise to {\em intrinsic noise}.

Performance comparisons based on a single run or a few runs do not take these noises
into account. Because we cannot assign the resulting statements a confidence measure, they
are more qualitative than quantitative. The degree to which
we can accept such statements depends on the noise level and more generally,
on the distribution of performance.

In this paper, we use recall as a performance measure (cf.~Section \ref{sec:expt}
and Section 3.2 in \cite{yeh-signif}).
Recall samples are obtained by resampling from training data and training classifiers 
on these samples.

The resampling methods used here are cross-validation and bootstrap
\cite[cf.~Section \ref{sec:bs}]{efron-gong-83,efron-tib-93}.
\ignore{
 \newcite{kohavi-95} studied the bias and variance of the two methods, in order to
find an optimal setup for model selection.} Section \ref{sec:exp}
presents the experimental goals and setup. Results are presented and discussed
in Section \ref{sec:res}, and a summary is provided in Section \ref{sec:sum}.

\section {The Bootstrap Method}
\label{sec:bs}

The bootstrap is a re-sampling technique designed for obtaining empirical distributions of
estimators. It can be thought of as a smoothed version of $k$-fold cross-validation (CV).
The method has been applied to decision tree and bayesian classifiers by \newcite{kohavi-95}
and to neural networks by, e.g., \newcite{lebaron-98}.

In this paper, we use the bootstrap method to obtain
the distribution of performance of a system which learns to identify non-recursive
noun-phrases (base-NPs). While there are a few refinements of the method,
the intention of this paper is to present the benefits of obtaining distributions,
rather than optimising bias or variance.
We do not aim to study the properties of bootstrap estimation.

Let a statistic $S=S(x_1,\dots,x_n)$ be a function of the independent observations
$\{x_i\}_{i=1}^n$ of a statistical variable $X$. The bootstrap method constructs the
distribution function of $S$ by successively re-sampling $x$ with replacements.

After $B$ samples, we have a set of {\em bootstrap samples}
$\{x_1^b,\dots,x_n^b\}_{b=1}^{B}$, each of which yields an estimate $\hat S^b$ for $S$.
The distribution of $\hat S$ is the bootstrap estimate for the distribution of
$S$. That distribution is mostly used for estimating the standard deviation, bias,
or confidence interval of $S$.

In the present work, $x_i$ are the base-NP instances in a given corpus,
and the statistic $S$ is the recall on a test set.

\section {Experimental Setup}
\label{sec:exp}

The aim of our experiments is to test whether the recall distribution can
be helpful in answering the questions Q1--Q3 mentioned in the introduction
of this paper.

The data and learning algorithms are presented in Sections \ref{sec:data} and
\ref{sec:lrn}. Section \ref{sec:pol} describes the sampling method in detail.
Section \ref{sec:expt} motivates the use of recall and describes the experiments.

\subsection{Data}
\label{sec:data}

We used Penn-Treebank \cite{TB-corpus} data, presented in Table \ref{tab:data}.
Wall-Street Journal (WSJ) Sections 15-18 and 20 were used by \newcite{ram-mar-95}
as training and test data respectively for evaluating their base-NP chunker. These data have
since become a standard for evaluating base-NP systems.

\begin{table}
\ifnum \toicml=1
\caption{Data sources}
\vskip 0.15in
\fi
\begin{tabular}{|l|c|c|c|}
\hline
 Source & Sentences & Words & Base \\
        &           &       & NPs \\
\hline
 WSJ 15-18 &  8936  & 229598 & 54760 \\
 WSJ 20    &  2012  &  51401 & 12335 \\
 ATIS      &   190  &  2046  & 613 \\
\hline
 WSJ 20a   &   100  &  2479  & 614 \\
 WSJ 20b   &    93  &  2661  & 619 \\
\hline
\end{tabular}
\ifnum \toacl=1
\caption{Data sources}
\fi
\label{tab:data}
\end{table}

The WSJ texts are economic newspaper reports, which often include elaborated
sentences containing about six base-NPs on the average.

The ATIS data, on the
other hand, are a collection of customer requests related to flight schedules.
These typically include short sentences which contain only three base-NPs on the average.
For example:
\begin{verbatim}
 I have a friend living in Denver
 that would like to visit me
 here in Washington DC .
\end{verbatim}

The structure of sentences in the ATIS data differs significantly from that in the WSJ data.
We expect this difference to be reflected in the recall of systems tested on both data sets.

The small size of the ATIS data can influence the results as well. To distinguish
the size effect from the structural differences, we drew two equally small
samples from WSJ Section 20. These samples, WSJ20a and WSJ20b, consist of the first
100 and the following 93 sentences respectively. There is a slight difference in size because sentences
were kept complete, as explained Section \ref{sec:pol}.

\subsection{Learning Algorithms}
\label{sec:lrn}

We evaluated base-NP learning systems based on two algorithms:
MBSL \cite{ADK99} and SNoW \cite{MPRZ99}.

MBSL is a memory-based system which records, for each POS sequence containing a border (left,
right, or both) of a base-NP, the number of times it appears with that border vs.~the number
of times it appears without it. It is possible to set an upper limit on the length of the POS
sequences.

Given a sentence, represented by a sequence of POS tags, the system examines each sub-sequence
for being a base-NP. This is done by attempting to tile it using POS sequences
that appeared in the training data with the base-NP borders at the same locations.

For the purpose of the present work, suffice it to mention that one of the
parameters is the {\em context size}$\,(c)$.
It denotes the maximal number of words considered before or after a base-NP when
recording sub-sequences containing a border.

SNoW \cite[``Sparse Network of Winnow'']{roth-snow} is a network architecture of Winnow
classifiers \cite{Litt88}. Winnow is a mistake-driven algorithm for learning a linear
separator, in which feature weights are updated by multiplication.
The Winnow algorithm is known for being able to learn well even in the presence
of many noisy features.

The features consist of one to four consecutive POSs in
a 3-word window around each POS. Each word is classified as a beginning of a base-NP,
as an end, or neither.

\subsection{Sampling Method}
\label{sec:pol}

In generating the training samples we sampled {\em complete} sentences.
In MBSL, an un-marked boundary may be counted as a negative example for the
POS-subsequences which contains it. Therefore, sampling only part of the base-NPs in
a sentence will generate negative examples.

For SNoW, each word is an example, but most of the words are neither a beginning nor
an end of a base-NP. Random sampling of words might generate a sample with
an improper balance between the three classes.

To avoid these problems, we sampled full sentences instead of
words or instances. Within a good approximation, it can be assumed that base-NP
patterns in a sentence do not correlate. The base-NP instances drawn from the
sampled sentences can therefore be regarded as independent.

As described at the end of Sec.~\ref{sec:data}, the WSJ20a and WSJ20b data were created
so that they contain 613 instances, like the ATIS data. In practice, the number of instances
exceeds 613 slightly
due to the full-sentence constraint. For the purpose of this work, it is enough that their
size is very close to the size of ATIS.

\begin{table}[bht]
\begin{tabular}{|l|c|c|}
\hline
 \multicolumn{1}{|c|}{Dataset} & Sentences &  Base-NPs \\
\hline
 Training  &  8938 $\pm$ 48  &  54763 $\pm$ 2 \\
 \quad Unique: & 5648 $\pm$ 34  & \\
\hline
\end{tabular}
\ifnum \toacl=1
\caption{Sentence and instant counts for the bootstrap samples. The second line
refers to unique sentences in the training data.}
\fi
\label{tab:bd}
\end{table}

We used the WSJ15-18 dataset for training. This dataset contains $n_0=54760$ base-NP instances.
The number of instances in a bootstrap sample depends on the number of instances in the
last sampled sentence. As Table \ref{tab:bd} shows, it is slightly more than $n_0$.

For $k$-CV sampling, the data were divided into $k$ random distinct
parts, each containing ${n_0 \over k} \pm 2$ instances. 
Table \ref{tab:expt} shows the number of recall samples in each experiment (MBSL and SNoW
experiments were carried out seperately).

\begin{table}[bht]
\begin{tabular}{|l|c|c|}
\hline
\multicolumn{1}{|c|}{Method} & MBSL &  SNoW \\
\hline
 Bootstrap & 2200 & 1000 \\
 CV (total folds) & 1500 & 1000 \\
\hline
\end{tabular}
\caption{Number of bootstrap samples and total CV folds.}
\label{tab:expt}
\end{table}

\subsection{Experiments}
\label{sec:expt}

 We trained SNoW and MBSL; the latter using context
sizes of $c$=1 and $c$=3. Data sets WSJ20, ATIS, WSJ20a, and WSJ20b were used for testing.
MBSL runs with the two $c$ values were conducted on the same training samples, therefore
it is possible to compare their results directly.

Each run yielded recall and precision. Recall may be viewed as the expected 0-1 loss-function
on the {\em given} test sample of instances. Precision, on the other hand, may be viewed as 
the expected 0-1 loss on the sample of instances {\em detected} by the learning
system. Care should be taken when discussing the distribution of precision values
because this sample varies from run to run.
We will therefore only analyse the distribution of recall in this work.

In the following, $r^1$ and $r^3$ denote recall samples of
MBSL with $c=1$ and $c=3$, with standard deviations $\sigma^1$ and $\sigma^3$.
$\rho^{13}$ denotes the cross-correlation between $r^1$ and $r^3$.
SNoW recall and standard deviation will be denoted by $r^\sn$ and $\sigma^\sn$. 

To approach the questions raised in the introduction we made the
following measurements: 

{\bf Q1:} System comparison was addressed by comparing $r^1$ and $r^3$
on the same test data. With samples at hand, we obtained an estimate of
$P(r^3 > r^1)$.

{\bf Q2:} We studied training and test adequacy through the effect of more specific
features on recall, and on its standard deviation.

Setting $c=3$ takes into account sequences with context of two and three words
in addition to those with $c=1$. Sequences with larger context are more specific,
and an improvement in recall implies that they are informative in the test data as well.

For particular choices of parameters and test data, the recall spread yields
an estimate of the training sampling noise.
On inadequate data, where the statistics differ significantly from those in
the training data, even small changes in the model can lead to a noticeable difference in
recall. This is because the model relies on statistics which appear relatively rarely
in the test data. Not only do these statistics provide little information about the
problem, but even small differences in weighting them are relatively influential.

Therefore, the more training and test data differ from each other, the more spread
we can expect in results.

{\bf Q3:} For comparing test data sets with a system, we used cross-correlations between $r^1$, $r^3$, or $r^\sn$
samples obtained on these data sets. We know that WSJ data
are different from ATIS data, and so expect the results on WSJ to correlate with ATIS
results less than with other WSJ results.


%

\section {Results and Discussion}
\label{sec:res}

For each of the five test datasets, Table \ref{tab:meanvar} reports averages and standard
deviations of $r^1$, $r^3$, and $r^\sn$ obtained by 3, 5, 10, and 20-fold cross-validation,
and by bootstrap. $\rho^{13}$ and $P(r^3>r^1)$ are reported as well.

We discuss our results by considering to what extent they
provide information for answering the three questions:

\paragraph{Q1 -- Comparing systems on given data:}
For the WSJ data sets, the difference between $r^3$ and $r^1$ was well above their
standard deviations, and $r^3>r^1$ nearly always. For ATIS, the standard deviation
of the difference
($\sigma^2_{r^3-r^1} = (\sigma^1)^2+(\sigma^3)^2-2\sigma^1\sigma^3\cdot\rho^{13}$)
 was small due to the high $\rho^{13}$, and $r^1>r^3$ nearly always.

\paragraph{Q2 -- The adequacy of training and test sets:}
It is clear that adding more specific features, by increasing the context,
improved recall on the WSJ test data and degraded it on the ATIS data.
This is likely to be an indication of the difference in syntactic structure between ATIS
and WSJ texts.

Another evidence of structural difference comes from standard deviations.
The spread of the ATIS results always exceeded that of the WSJ results, with {\em all three
experiments}. That difference cannot be solely attributed to the small size of ATIS, since
WSJ20a and WSJ20b results displayed a much smaller spread. Indeed, these results had
a wider standard deviation than WSJ20, probably due to the smaller size, but not as
wide as ATIS. This indicates that base-NPs in ATIS text have different characteristics than
those in WSJ texts.

\paragraph{Q3 -- Comparing datasets by a system:}
Table \ref{tab:corr} reports, for each pair of datasets, the correlation between the 5-fold CV
recall samples of each experiment on these datasets. The correlations change with CV fold
number, 5-fold results were chosen as they represent intermediary values.

Both MBSL experiments yielded negligible correlations of ATIS results with
{\em any} WSJ data set, whether large or small. These correlations were always
weaker than with WSJ20a and WSJ20b, which are about the same size.

This is due to ATIS being a different kind of text. The correlation between
WSJ20a and WSJ20b results was also weak. This may be due to their small sizes; these texts
might not share enough features to make a significant correlation.

SNoW results were highly correlated for all pairs.
That behaviour is markedly different from the MBSL results, and indicates a
high level of noise in the SNoW features. Indeed, Winnow is able to learn well in the
presence of noise, but that noise causes the high correlations observed here.

\subsection{Further Observations}

The decrease of $\rho^{13}$ with CV fold number is related to
stabilization of the system. As the folds become larger, training samples become more similar
to each other, and the spread of results decreases. This effect was not visible in
the SNoW data, most likely due to the high level of noise in the features. This noise also
contributes to the higher standard deviation of SNoW results.

\begin{table*}[p]
\begin{tabular}{|l|l|c|c|c|c|c|}
\hline
  Test & Method & \multicolumn{4}{c|}{MBSL} & SNoW \\
\cline{3-7}
  data & & $E(r^1)\pm \sigma^1$ & $E(r^3)\pm \sigma^3$
                & $\rho^{13}$ & $P(r^3>r^1)$ & $E(r^\sn)\pm \sigma^\sn$ \\
\hline
\ignore{ 
 & 3-CV    & 89.83 $\pm$ 0.24 & 91.53 $\pm$ 0.22 & 0.70 & 100\% & 90.22 $\pm$ 1.05 \\
 & 5-CV    & 89.94 $\pm$ 0.31 & 91.70 $\pm$ 0.30 & 0.77 & 100\% & 90.40 $\pm$ 1.13 \\
 Random & 10-CV   & 90.02 $\pm$ 0.46 & 91.80 $\pm$ 0.45 & 0.80 & 100\% & 90.52 $\pm$ 1.29 \\
 test   & 20-CV   & 90.04 $\pm$ 0.63 & 91.84 $\pm$ 0.62 & 0.80 & 100\% & 90.53 $\pm$ 1.44 \\
\cline{2-7}
 & Bootstrap      & 89.76 $\pm$ 0.24 & 91.43 $\pm$ 0.21 & 0.69 & 100\% & 89.91 $\pm$ 0.95 \\
\cline{2-7}
 & $E(\cdot)$ & 93.06 & 96.97 &  \multicolumn{2}{c|}{}  & 95.83 \\
\hline }
 & 3-CV    & 89.64 $\pm$ 0.16 & 91.26 $\pm$ 0.12 & 0.36 & 100\% & 90.18 $\pm$ 1.01 \\
 & 5-CV    & 89.75 $\pm$ 0.14 & 91.43 $\pm$ 0.10 & 0.30 & 100\% & 90.37 $\pm$ 1.03 \\
 WSJ 20 & 10-CV   & 89.80 $\pm$ 0.12 & 91.53 $\pm$ 0.08 & 0.25 & 100\% & 90.47 $\pm$ 1.11 \\
 & 20-CV   & 89.81 $\pm$ 0.11 & 91.56 $\pm$ 0.07 & 0.28 &  100\% &90.51 $\pm$ 1.19 \\
\cline{2-7}
 & Bootstrap      & 89.58 $\pm$ 0.17 & 91.16 $\pm$ 0.14 & 0.42 & 100\% & 89.83 $\pm$ 0.93 \\
\cline{2-7}
 & $E(\cdot)$ & 89.74 & 91.58 & \multicolumn{2}{c|}{} & 91.23 \\
\hline
 & 3-CV    & 85.70 $\pm$ 2.03 & 83.99 $\pm$ 1.87 & 0.82 & 3\% & 83.70 $\pm$ 4.11 \\
 & 5-CV    & 85.76 $\pm$ 1.87 & 83.69 $\pm$ 1.57 & 0.79 & 1\% & 83.53 $\pm$ 4.52 \\
ATIS & 10-CV   & 85.90 $\pm$ 1.31 & 84.78 $\pm$ 0.92 & 0.78 & 4\% & 83.38 $\pm$ 5.14 \\
 & 20-CV   & 85.78 $\pm$ 1.16 & 83.28 $\pm$ 0.85 & 0.77 & 0\% & 83.23 $\pm$ 5.36 \\
\cline{2-7}
 & Bootstrap      & 85.72 $\pm$ 1.95 & 84.69 $\pm$ 1.95 & 0.81 & 16\% & 83.50 $\pm$ 3.35 \\
\cline{2-7}
 & $E(\cdot)$ & 85.81 & 83.20 &  \multicolumn{2}{c|}{} & 85.48 \\
\hline
 & 3-CV    & 89.45 $\pm$ 0.42 & 91.25 $\pm$ 0.56 & 0.33 & 100\% & 90.84 $\pm$ 1.04 \\
 & 5-CV    & 89.66 $\pm$ 0.36 & 91.64 $\pm$ 0.54 & 0.32 & 100\% & 91.07 $\pm$ 1.15 \\
 WSJ 20a & 10-CV   & 89.79 $\pm$ 0.28 & 91.85 $\pm$ 0.49 & 0.20 & 100\% & 91.14 $\pm$ 1.26 \\
 & 20-CV   & 89.82 $\pm$ 0.23 & 91.89 $\pm$ 0.44 & 0.18 &  100\% &91.11 $\pm$ 1.39 \\
\cline{2-7}
 & Bootstrap      & 89.42 $\pm$ 0.47 & 91.55 $\pm$ 0.57 & 0.33 & 99\% & 90.76 $\pm$ 1.00 \\
\cline{2-7}
 & $E(\cdot)$ & 89.73 & 92.18 &  \multicolumn{2}{c|}{}  & 90.07 \\
\hline
 & 3-CV    & 88.95 $\pm$ 0.41 & 90.12 $\pm$ 0.39 & 0.37 & 99\% & 89.79 $\pm$ 0.81 \\
 & 5-CV    & 89.03 $\pm$ 0.36 & 90.15 $\pm$ 0.31 & 0.31 & 99\% & 89.81 $\pm$ 0.84 \\
WSJ 20b & 10-CV   & 89.06 $\pm$ 0.33 & 90.14 $\pm$ 0.22 & 0.28 & 99\% & 89.83 $\pm$ 0.86 \\
 & 20-CV   & 89.07 $\pm$ 0.27 & 90.13 $\pm$ 0.18 & 0.22 & 100\% & 89.87 $\pm$ 0.88 \\
\cline{2-7}
 & Bootstrap      & 89.00 $\pm$ 0.44 & 90.17 $\pm$ 0.44 & 0.38 & 98\% & 89.93 $\pm$ 0.80 \\
\cline{2-7}
 & $E(\cdot)$ & 89.01 & 91.55 & \multicolumn{2}{c|}{}  & 90.79 \\
\hline
\end{tabular}
\caption{Recall statistic summary for MBSL with contexts $c=1$ and $c=3$, and SNoW. The $E(\cdot)$
figures were obtained using the full training set.
Note the monotonic change of standard deviation with fold number. The s.d. of the bootstrap samples are
closest to those of low-fold CV samples.}
\label{tab:meanvar}
\end{table*}

\begin{table*}[p]
\ignore{ 
\begin{tabular}{|l|c|c|c|c|c|c|c|c|c|c|c|c|}
\hline
 5-CV   & \multicolumn{3}{c|}{WSJ 20b} & \multicolumn{3}{c|}{WSJ 20a}
        & \multicolumn{3}{c|}{ATIS}& \multicolumn{3}{c|}{WSJ 20} \\
    &  $r^1$ & $r^3$ & $r^\sn$ &  $r^1$ & $r^3$ & $r^\sn$
    &  $r^1$ & $r^3$ & $r^\sn$ &  $r^1$ & $r^3$ & $r^\sn$ \\
\hline
Random  & 0.02 & -0.05 & 0.68 & 0.09 & 0.00 & 0.73 & 0.04 & 0.00 & 0.75 & 0.22 & 0.03 & 0.95 \\
 \cline{11-13}
 WSJ 20  & 0.33 & 0.19 & 0.72 & 0.26 & 0.29 & 0.78 & 0.08 & 0.02 & 0.76 & \multicolumn{3}{l}{}  \\
 \cline{8-10}
 ATIS    & -0.01 & 0.00 & 0.59 & 0.02 & -0.01 & 0.63 &\multicolumn{6}{l}{}  \\
 \cline{5-7}
 WSJ 20a & 0.07 & 0.04 & 0.59 &\multicolumn{9}{l}{} \\
 \cline{1-4}
\end{tabular} }
\begin{tabular}{|l|c|c|c|c|c|c|c|c|c|}
\hline
 5-CV   & \multicolumn{3}{c|}{WSJ 20b} & \multicolumn{3}{c|}{WSJ 20a}
        & \multicolumn{3}{c|}{ATIS} \\
    &  $r^1$ & $r^3$ & $r^\sn$ &  $r^1$ & $r^3$ & $r^\sn$
    &  $r^1$ & $r^3$ & $r^\sn$ \\
\hline
 WSJ 20  & 0.33 & 0.19 & 0.72 & 0.26 & 0.29 & 0.78 & 0.08 & 0.02 & 0.76   \\
 \cline{8-10}
 ATIS    & -0.01 & 0.00 & 0.59 & 0.02 & -0.01 & 0.63 &\multicolumn{3}{l}{}  \\
 \cline{5-7}
 WSJ 20a & 0.07 & 0.04 & 0.59 &\multicolumn{6}{l}{} \\
 \cline{1-4}
\end{tabular}
\caption{Cross-correlations between recalls of the three experiments on the test data for five-fold
CV. Correlations of $r^1$ capture dataset similarity in the best way.}
\label{tab:corr}
\end{table*}

\section{Summary and Further Research}
\label{sec:sum}

In this work, we used the distribution of recall to address questions concerning
base-NP learning systems and corpora. Two of these questions, of training and test
adequacy, and of comparing data sets using NLP systems, were not addressed before.

The recall distributions
were obtained  using CV and bootstrap resampling.

We found differences between algorithms with similar recall, related to the
 features they use.

We demonstrated that using an inadequate test set may lead to noisy performance results.
This effect was observed with two
different learning algorithms. We also reported a case when changing a parameter of
a learning algorithm improved results on one dataset but degraded results on another.

We used classifiers as ``similarity rulers'', for producing a similarity measure
between datasets.
Classifiers may have various properties as similarity rulers, even when their
recalls are similar. Each classifier should be
scaled differently according to its noise level.
This demonstrates the way we can use classifiers to study data, as well as
use data to study classifiers.

By using MBSL with different context sizes, our results provide insights into the
relation between training and test data sets,
in terms of general and specific features.  That issue becomes important when one plans
to use a system trained on certain data set for analysing an arbitrary text. Another approach
to this topic, examining the effect of using lexical bigram information, which is very
corpus-specific, appears in \cite{gildea-emnlp01}.

In our experiments with systems trained on WSJ data, there was a clear difference between
their behaviour on other WSJ data and on the ATIS data set, in which the structure of
base-NPs is different. That difference
was observed with correlations and standard deviations. This shows that resampling
the training data is essential for noticing these structure differences.

To control the effect of small size of the ATIS dataset, we provided
two equally-small WSJ data sets. The effect of different genres was stronger
than that of the small-size.

In future study, it would be helpful to study the distribution of recall using
training and test data from a few genres, across genres, and on combinations
(e.g.~``known-similarity corpora'' \cite{kilgarriff-emnlp3}). This will provide
a measure of the transferability of a model.

\ignore{
In order to further study the methods employed in this work,
we would like to invoke them on data from other genres, and to use
other systems. In some cases, it may be possible to provide approximate
analytic descriptions of the recall distributions.}

We would like to study whether there is a relation between bootstrap and 2 or
3-CV results. The average number of unique base-NPs in a random bootstrap training
sample is about 63\% of the total training instances (Table \ref{tab:bd}). That
corresponds roughly to the size of a 3-CV training sample. More work is required to
see whether this relation between bootstrap and low-fold CV is meaningful.

We also plan to study the distribution of precision. As mentioned in
Sec.~\ref{sec:expt}, the precisions of different runs are now taken from different
sample spaces. This makes the bootstrap estimator unsuitable, and more study
is required to overcome this problem.

\ifnum \toacl=1
\bibliographystyle{acl}
\fi

\ifnum \toicml=1
\bibliographystyle{mlapa}
\fi
\bibliography{paper}

\end{document}